\documentclass[letterpaper]{article}

\usepackage[preprint]{aaai2027}
\usepackage[hyphens]{url}
\usepackage{graphicx}
\usepackage{natbib}
\usepackage{caption}
\usepackage{amsmath,amssymb}
\usepackage{booktabs,tabularx}
\usepackage{listings}
\usepackage{xcolor}

\makeatletter
\newcommand{\inputwidetable}[1]{%
    \begingroup
    \def\table{\@ifnextchar[{\cae@widetable}{\cae@widetable[]}}%
    \expandafter\let\expandafter\endtable\csname endtable*\endcsname
    \input{#1}%
    \endgroup
}
\def\cae@widetable[#1]{\csname table*\endcsname[!tp]}
\makeatother

\definecolor{darkgreen}{rgb}{0,0.5,0}

\title{CAE-Bench: Benchmarking AI Agents for Reusable CAE Solver Development}

\author{
    Zhiyu Zhou\textsuperscript{\rm 1}\equalcontrib,
    Tianyou Liu\textsuperscript{\rm 1}\equalcontrib,
    Xinping Yang\textsuperscript{\rm 3}\equalcontrib,
    Yiyao Liu\textsuperscript{\rm 1,2},
    Jiayi Dong\textsuperscript{\rm 1,2},\\
    Yihan Lin\textsuperscript{\rm 2},
    Haokai Ye\textsuperscript{\rm 2},
    Yifan Lu\textsuperscript{\rm 2},
    Zeyu Dong,
    Yansong Li\textsuperscript{\rm 2}\corresponding,
    Shengqi Zhang\textsuperscript{\rm 2}
}
\affiliations{
    \textsuperscript{\rm 1}Southern University of Science and Technology\\
    \textsuperscript{\rm 2}Eastern Institute of Technology, Ningbo\\
    \textsuperscript{\rm 3}Peking University\\
    ysli@eitech.edu.cn
}

\begin{document}
\maketitle
\begin{abstract}

    We developed CAE-Bench, a benchmark to evaluate AI agents on Computer-Aided Engineering (CAE) solver development. A CAE solver is more than a PDE solver; it must be able to generalize across various inputs, such as different boundary conditions, geometries, and material properties. CAE-Bench evaluates the generalizability of AI-developed CAE solvers by providing the same task, such as a heat transfer problem, multiple cases, with each case specifying an input. The AI agent develops a solver for a fixed task using only a few public cases, and the resulting solver is then tested on held-out cases, i.e., cases it has never seen before. We evaluate 15 models across a range of reasoning settings on 80 computational tasks. Our experiments show that even if a solver developed by an AI agent performs well on the public cases, it may even get $0$ score on held-out cases. Even GPT-6.2 Astra, the highest-scoring model in our evaluation, averages 42.48 on held-out cases for tasks with varying physical inputs. A data format and an SDK for task registration are provided, so the researcher can use the benchmark on their own infrastructure. We hope that CAE-Bench will help the community to develop AI agents that can develop reusable CAE solvers that generalize to various inputs.

\end{abstract}
\section{Introduction} \label{sec:introduction}

For a PDE problem with given geometry and boundary conditions, it is not surprising that AI agents are capable of developing an accurate numerical solver. The AI-developed solvers for a fixed PDE problem can be evaluated through PDEAgent-Bench~\citep{hang2026pdeagent} or CFDLLMBench~\citep{somasekharan2026cfdllmbench}. However, for an industrial-level Computer-Aided Engineering (CAE) solver, solving a fixed PDE is not enough. The solver must be able to handle different geometries, boundary conditions, and material properties. The inputs of held-out cases should not be provided during development. Therefore, for a single \emph{task}, the AI agent is required to develop a CAE solver that can generalize across \emph{cases} with varying inputs. In this paper, we introduce \textbf{CAE-Bench}, a benchmark to evaluate the ability of AI agents to develop reusable CAE solvers that generalize across inputs.

We consider an extreme example. The task is to develop a solver for the governing equations of a 2-dimensional incompressible flow of a Newtonian fluid. The solver predicts velocity and pressure fields for given inputs. If the AI agent is only required to solve the task for fixed inputs, and the inputs are provided to the agent during development, then the agent can compute and store the answer during development, then return it during evaluation. This does not demonstrate that the agent can develop a reusable solver, since the resulting solver only works for these fixed inputs.

To evaluate the generalizability of AI-developed CAE solvers, we evaluate solvers on the same task with different cases, each case containing inputs and the reference used for scoring. All cases are distinguished into \emph{Public cases} and \emph{Held-out cases}. Public cases provide public inputs and example outputs during development. The example outputs have the required format, but their values are random. Held-out cases contain inputs that vary in geometry, material response, initial states, or boundary-condition types. These held-out cases are not provided to the agent during development. In summary, the public cases are used in both development and evaluation, while the held-out cases are only used in evaluation.

Besides input, each case is associated with reference results collected from analytical solutions, existing datasets, and established simulation software. These reference results are used for scoring during the evaluation phase, as discussed in Section~\ref{sec:task}.

We also designed a data format to store the requirements of a given task, cases, and the output. Details are discussed in Section~\ref{app:contract}. An SDK with tools for task registration and evaluation is also provided. Therefore, researchers can use the benchmark on their own infrastructure to evaluate the performance of their LLM model or harness.

The RC7 release of CAE-Bench contains 80 computational tasks, which include 13 prescribed-PDE tasks, 43 fluid-mechanics tasks, and 24 fixed-problem tasks adapted from CFDCodeBench~\citep{somasekharan2026cfdllmbench}. Fixed-problem tasks do not have held-out cases and are reported separately. Replication of the complete evaluation is discussed in Appendix~\ref{app:reference-records}.

Using the RC7 release, we evaluate 15 models across a range of reasoning settings. At medium effort, some AI-developed solvers receive full score on all public cases of a task. However, all AI-developed solvers got $0$ score on at least 1 held-out case of the same task.  Also, most 0-score cases reach numerical scoring, which means the solver runs successfully and the output format is correct. However, the output fails the accuracy requirements. These results show that the current best models are not capable of developing an industrial-level reusable CAE solver without human involvement.

\paragraph{Related work} \label{sec:related}

Several works have explored the ability of AI agents to develop numerical solvers. CodePDE~\citep{li2026codepde} uses debugging and iterative refinement to improve AI-developed PDE solvers, and tests them on batches of initial conditions. PDEAgent-Bench~\citep{hang2026pdeagent} evaluates solver generation from individual PDE specifications using libraries such as DOLFINx, Firedrake, and deal.II. CFDLLMBench~\citep{somasekharan2026cfdllmbench} evaluates AI agents on CFD-related tasks, including code generation and OpenFOAM workflows.

MetaOpenFOAM, Foam-Agent, OpenFOAMGPT 2.0, and ChatCFD automate CFD workflows from user instructions~\citep{chen2024metaopenfoam,yue2026foamagent,feng2026openfoamgpt,fan2026chatcfd}. \citet{xiao2026cfdagents} test general-purpose coding agents on CFD case setup and execution, including mesh generation. AI CFD Scientist uses OpenFOAM for parameter sweeps and changes to physical models~\citep{somasekharan2026cfdscientist}. For tasks with varying physical inputs, CAE-Bench tests a single submitted solver on held-out cases.

PDEBench, PDEArena, and The Well provide simulation data for evaluating learned PDE models~\citep{takamoto2022pdebench,gupta2023pdearena,ohana2024well}. PINNacle compares physics-informed neural networks~\citep{hao2024pinnacle}, and Poseidon tests pretrained PDE models on downstream problems~\citep{herde2024poseidon}.

SciCode tests code generation for scientific research problems~\citep{tian2024scicode}. SWE-bench and SWE-bench Science evaluate changes to existing software repositories~\citep{jimenez2024swebench,xu2026swebenchscience}, while Terminal-Bench evaluates agents on tasks in terminal environments~\citep{merrill2026terminalbench}.
AssetOpsBench evaluates agents on industrial asset operations and maintenance tasks in a simulated environment~\citep{patel2026assetopsbench}.

\section{Tasks and Evaluation Protocol} \label{sec:task}

For a benchmark framework, the most important aspect is the clear definition of tasks and evaluation protocols (scoring). The detailed definition of tasks and cases is discussed in Section~\ref{sec:problem-families} and the scoring rules are described in Section~\ref{sec:assessment}. Besides definitions, we discuss the input data format in Section~\ref{sec:contract} and the evaluation report format in Section~\ref{sec:assessment}.

\subsection{Tasks, Cases, and Observations} \label{sec:problem-families}

In this paper, the term \emph{task} defines a physical or mathematical problem. By define, we mean that the task specifies the required inputs, outputs, scoring rules and execution requirements. A single task contains one or more evaluation cases. The term \emph{case} defines a complete test example for the task. Each case contains specified inputs and reference results.
Formally:
\begin{equation} \label{eq:case-components}
    \text{case} = \text{inputs} + \text{reference result}.
\end{equation}

The AI-developed solver must handle all the cases assigned to the task. Table~\ref{tab:task-case} shows the task-case relationship in a released flow task.

\begin{table*}[!tp]
    \centering
    \small
    \begin{tabularx}{\linewidth}{@{}l>{\raggedright\arraybackslash}X>{\raggedright\arraybackslash}X@{}}
        \toprule
        Aspect       & Fixed by the task                                                                & Supplied by each case                                                             \\
        \midrule
        Analysis     & Steady, 2-dimensional, single-phase laminar flow; incompressible Newtonian fluid & Fluid domain; positive constant density and viscosity                             \\
        Conditions   & Supported mechanical boundary conditions, body forcing, and pressure conventions & Condition types, boundary assignments, values, and applicable pressure constraint \\
        Observations & Definitions and units of velocity, pressure, and other supported quantities      & Requested quantities, locations, and point, average, or integral observations     \\
        \bottomrule
    \end{tabularx}
    \caption{Cases for the 2-dimensional steady incompressible Newtonian-flow task.}
    \label{tab:task-case}
\end{table*}

The \emph{inputs} describe the physical conditions, such as geometry, material properties, and boundary or initial conditions. During evaluation, the submitted solver receives the inputs and produces its own outputs.

Evaluation cases can be distinguished into two groups: \emph{Public cases} and \emph{Held-out cases}. \emph{Public cases} provide inputs and template outputs (which means the data format is identical to the reference results but the numerical values are randomly generated) during development. \emph{Held-out cases} are not accessible during development but only used for evaluation. A task may contain either or both case groups.

In this paper, we call each requested output produced by an AI-developed solver for a given case an \emph{observation}. Each observation specifies a physical quantity, its units, and its form. An observation may contain values at multiple locations and times. For example, temperature is a physical quantity, while point values, cell averages, and spatial integrals are different forms of temperature observations. A temperature field can also be observed at a specific time or averaged over a period of time, resulting in different observations.

\begin{figure*}[!tp]
    \centering
    \includegraphics[width=\linewidth]{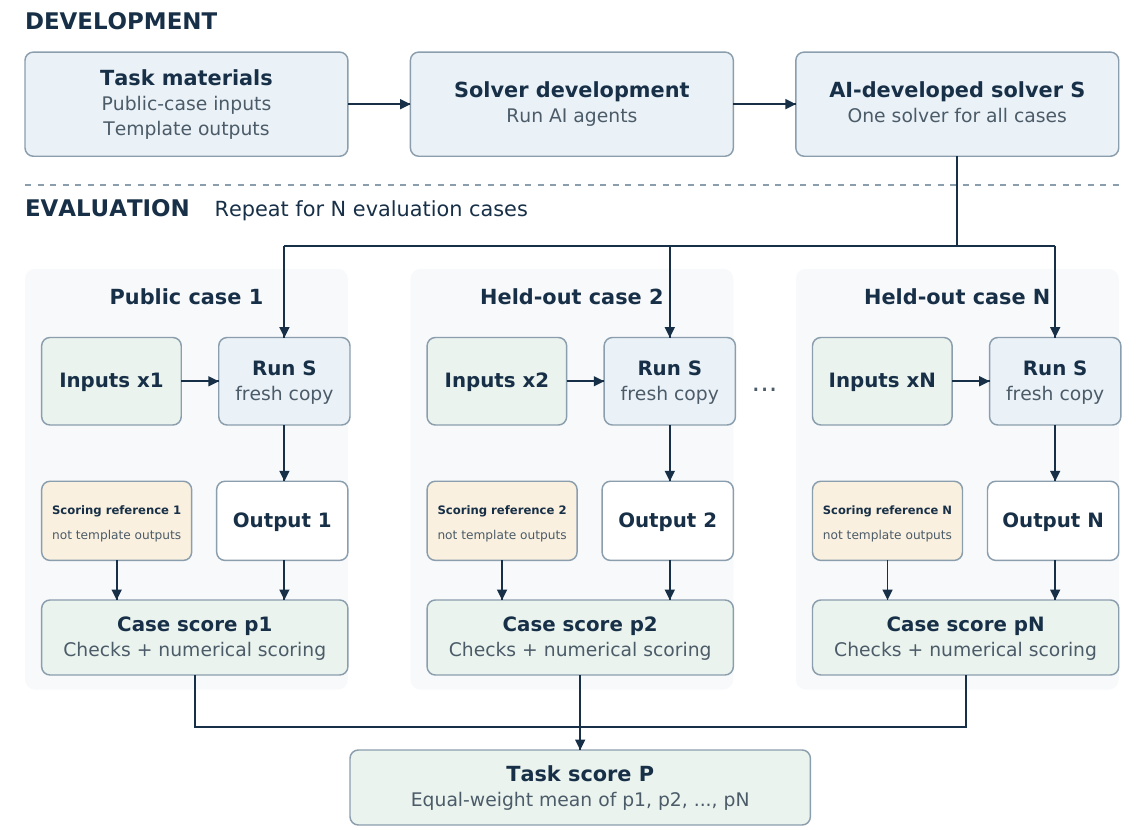}
    \caption{The workflow of CAE-Bench}
    \label{fig:workflow}
\end{figure*}

\subsection{Data Formats and SDK} \label{sec:contract}

Each task contains instructions, \texttt{DATA.md}, and the shared \texttt{CONTRACT.md} as the agent's initial prompt. \texttt{DATA.md} explains how to read the inputs and write the outputs. \texttt{CONTRACT.md} defines the common file formats.

In our implementation, an HDF5 file is used to store the sample identifiers, values, and field arrays. For sampling based on a mesh, the sample identifiers refer to node or element tags in a Gmsh~\citep{geuzaine2009gmsh} \texttt{.msh} file.
Each case input includes a \texttt{problem.json} file. \texttt{problem.json} specifies the physical setup, references the data files, and declares the requested outputs.

We also provide an optional SDK to help agents develop without having trouble understanding the data format. The SDK includes tools that test whether files meet format and compatibility requirements without providing evidence that shows numerical accuracy.
Detailed discussions of the data format and the SDK are included in Appendix~\ref{app:contract}.

These given prompts and tools guide the agent to develop solvers that follow the task's physical assumptions and restrictions. Different tasks have different restrictions. For example, some tasks may require the solver to solve prescribed equations while other tasks let the agent decide the equations to solve.

The agent can use analytical expressions, numerical libraries, existing open-source software, or combine them, to develop a CAE solver.  Therefore, CAE-Bench does not require Agents to implement every numerical method from scratch.

The submitted AI-developed CAE solver must include a \texttt{solve.py} entry point. CAE-Bench evaluates through the entry point for each case by
\begin{lstlisting}
python solve.py --input problem.json \
    --output result.h5
\end{lstlisting}

The submitted AI-developed solver is fixed after the development phase. The agent cannot revise the solver during the evaluation phase. For each case, CAE-Bench makes a fresh writable copy of the saved solver. The running solver receives the current case input. For that case, the solver must give the requested output. During evaluation, network access is not allowed. The solver must finish the case within the declared time, CPU, memory, and accelerator limits. Figure~\ref{fig:workflow} shows the entire workflow of CAE-Bench including development, execution, and evaluation phases.

\subsection{Scoring Rules and Evaluation Report} \label{sec:assessment}

Before giving any score, we first check the execution and output validity of AI-developed solvers. The solver receives $0$ score in one case if it exceeds a time or resource limit, exits with an error, fails to produce output, or produces invalid output. CAE-Bench checks output identifiers, time axes, array shapes, and data types against the requirements. The output values must be finite and data types such as integer and Boolean must follow the exact agreement as in template outputs. Failure of a required type check sets the score to $0$.

For floating-point observations, we compare the solver's outputs with the reference values at all requested times and locations, including initial frames and boundaries. To calculate the score, we divide the outputs into groups based on the physical quantity, its components, and time. Each group contains all spatial values of one component at a given time, which are compared with the corresponding reference values.

For example, consider a case that requires 2-dimensional velocity and pressure fields at $10$ different times. Since velocity has $2$ components and pressure has $1$ component, we have $20$ groups for velocity and $10$ groups for pressure, resulting in $30$ groups in total. The number of spatial locations does not affect the number of groups, but only the number of values compared within each group.

We use a weighted normalized root-mean-square error (NRMSE) to evaluate each group. Let $M$ be the number of floating-point groups in a case, and let $i$ index the spatial locations within group $j$. We use $\widehat z_{ji}$ and $z_{ji}$ to denote the solver's output and reference value, respectively. The NRMSE for group $j$ is defined as
\begin{equation}
    e_j =
    \frac{\displaystyle
        \sqrt{\frac{\sum_i w_{ji}(\widehat z_{ji}-z_{ji})^2}{\sum_i w_{ji}}}}
    {\displaystyle
        \max\!\left(
        \sqrt{\frac{\sum_i w_{ji}(z_{ji}-b_j)^2}{\sum_i w_{ji}}},\ s_j
        \right)}.
    \label{eq}
\end{equation}
The default values of spatial weights $w_{ji}>0$ are equal for all $i$ and $j$, and the default value for $b_j$ is $0$.
The scale floor $s_j>0$ prevents division by zero. Spatial weights, scale floors, and scoring transformations are specified in the task descriptions and applied during evaluation. No further alignment or rescaling is allowed.

We denote the RMS of the $M$ group errors as $E$ and define the function $p(E)$ that maps $E$ to a dimensionless score between 0 and 1. Formally,
\begin{equation}
    \begin{aligned}
    E &= \sqrt{\frac{1}{M}\sum_{j=1}^{M} e_j^2},\\
    p(E)&=
    \begin{cases}
        1,             & E\leq 0.05,  \\
        (0.10-E)/0.05, & 0.05<E<0.10, \\
        0,\:           & E\geq 0.10.
    \end{cases}
    \end{aligned}
    \label{eq:case-score}
\end{equation}

Let $\mathcal T$ denote a task and $N_{\mathcal T}$ be the number of cases in task $\mathcal T$. Index $k$ runs from 1 to $N_{\mathcal T}$ and identifies a case. The score of case $k$ is denoted by $p_k$. The case score $p_k = 0$ if any required check fails. Otherwise, $p_k=p(E)$.

The task score $P_{\mathcal T}$ is defined as the mean of all case scores (all assigned cases contribute, whether they are public or held-out cases). Formally,
\begin{equation}
    P_{\mathcal T} = \frac{1}{N_{\mathcal T}}
    \sum_{k=1}^{N_{\mathcal T}}p_k.
    \label{eq:task-score}
\end{equation}

For tasks with varying physical inputs, we report scores on public cases and held-out cases separately to evaluate how well the solver generalizes to unseen inputs. For fixed-problem tasks (tasks do not contain held-out inputs), the scores only reflect performance on the given inputs. Infrastructure tests are evaluated using exact-match or pass/fail checks, and their results are reported separately from computational tasks.

Researchers can use adapters we provided to run the evaluation tools in their own environments.

\section{Benchmark Construction} \label{sec:construction}

In this section, we discuss how to construct tasks and cases and how to obtain their reference results. Before discussing the details, we summarize the number of tasks and cases in CAE-Bench (RC7) in Section~\ref{sec:benchmark-scope}. The design of cases for each task is described in Section~\ref{sec:case-coverage}. How the reference results are obtained is discussed in Section~\ref{sec:reference-construction}.

\subsection{Benchmark Contents} \label{sec:benchmark-scope}

As shown in Table~\ref{tab:benchmark-composition}, CAE-Bench (RC7) contains 2 infrastructure tests and 80 computational tasks. The 80 computational tasks are divided into 3 categories: prescribed-PDE tasks, fluid-mechanics tasks, and fixed-problem tasks.

The prescribed-PDE tasks cover elliptic problems, transport, reaction and diffusion, and coupled systems. The fluid-mechanics tasks cover single-phase flow, energy transport, turbulence, interfaces, thin films, shallow water, rarefied gases, and fluids with memory.
Both categories contain tasks with varying physical inputs.

The fixed-problem category uses 24 problems from CFDCodeBench~\citep{somasekharan2026cfdllmbench}.  Fixed-problem tasks do not have held-out cases.

\begin{table}[!htbp]
    \centering
    \small
    \begin{tabularx}{\linewidth}{@{}>{\raggedright\arraybackslash}Xrrrr@{}}
        \toprule
        Task category        & Tasks & \shortstack{Evaluation\\cases} & Public & \shortstack{Held\\out} \\
        \midrule
        Prescribed PDEs      & 13    & 266              & 3      & 263      \\
        Fluid mechanics      & 43    & 340              & 43     & 297      \\
        Fixed-problem tasks  & 24    & 24               & 24     & 0        \\
        \midrule
        Computational total  & 80    & 630              & 70     & 560      \\
        Infrastructure tests & 2     & 17               & 1      & 16       \\
        \bottomrule
    \end{tabularx}
    \caption{Number of tasks and cases of CAE-Bench (RC7).}
    \label{tab:benchmark-composition}
\end{table}

The total 630 computational cases comprise 70 public cases and 560 held-out cases. All 24 cases in the fixed-problem category are public cases. The 2 infrastructure are used to check data handling and scoring behavior. In the final report, we exclude these infrastructure tests.

\subsection{Case Design and Coverage} \label{sec:case-coverage}

For the same task, we create various cases by changing material properties, geometry, boundary conditions, and initial conditions.

For example, the steady heat-conduction task (which is included in RC7) contains 32 cases with different geometries, thermal conductivities, heat sources, and boundary conditions. The boundary conditions include Dirichlet, Neumann, Robin, and mixed conditions. Among these cases, 4 use pure Neumann conditions with an additional mean-temperature constraint.

Figure~\ref{fig:case-variation} shows the input variations for another 2 tasks. The periodic advection task contains 24 cases, combining 8 velocities with 3 initial fields. The generalized-Newtonian flow task contains 8 cases with different material models and conditions. The material models include power-law, Cross, Carreau, and Carreau--Yasuda. Another task, transient viscoelastic flow, contains 9 cases with different initial conformation and wall-driving histories, including stopping and reversing wall motion.

\subsection{Reference Results} \label{sec:reference-construction}

For each case, we get reference results from analytical solutions, manufactured solutions~\citep{roache2002manufactured}, existing simulation datasets, and existing simulation software. Table~\ref{tab:reference-evidence} shows how these reference results are collected for each task category.

\begin{table*}[!tp]
    \centering
    \small
    \begin{tabularx}{\linewidth}{@{}>{\raggedright\arraybackslash}p{0.20\linewidth}>{\raggedright\arraybackslash}p{0.30\linewidth}>{\raggedright\arraybackslash}X@{}}
        \toprule
        Task category                                                                & Source & Reference preparation \\
        \midrule
        Heat conduction\newline(2 tasks, 56 cases)                                   &
        Manufactured solutions and analytical series                                 &
        Evaluate the prescribed fields at the requested nodes and times.                                              \\
        \addlinespace
        PDEBench adaptations\newline(11 tasks, 210 cases)                            &
        Source trajectories and recomputations described in the accompanying records &
        Match the equations, initial states, time coordinates, units, and observations to the task.                   \\
        \addlinespace
        Basic steady flow\newline(2 tasks, 20 cases)                                 &
        Analytical solutions, Lethe, and Elmer                                       &
        Evaluate analytical fields or convert numerical solutions into the requested observations.                    \\
        \addlinespace
        Other fluid-mechanics tasks\newline(41 tasks, 320 cases)                     &
        OpenFOAM, Elmer, SPARTA, PyClaw, Thetis, and formula evaluation              &
        Evaluate formulas or convert solver fields into the requested values, integrals, or statistics.               \\
        \addlinespace
        Fixed-problem adaptations\newline(24 tasks, 24 cases)                        &
        Analytical representations, NGSolve, Dedalus, and OpenFOAM                   &
        Apply the problem definitions described in the accompanying records. Export the requested observations.       \\
        \bottomrule
    \end{tabularx}
    \caption{Reference sources and preparation for the computational suite.}
    \label{tab:reference-evidence}
\end{table*}

The basic steady-flow cases use both analytical and numerical reference results. 8 basic 2-dimensional steady-flow cases use analytical solutions. 2 other basic 2-dimensional steady-flow cases use Lethe calculations. For the basic 3-dimensional steady-flow cases, we generate reference results with Elmer.

For the tasks borrowed from PDEBench~\citep{takamoto2022pdebench}, we use existing simulation datasets or recompute the solutions when necessary. For example, the Darcy task requires a steady solution, but the original dataset only provides states at finite times, so it requires a recomputation.

For the fixed-problem tasks adapted from CFDCodeBench, we compute reference results using the specified equations and boundary conditions. For example, in the cavity cases, the velocity boundary conditions are prescribed. The momentum and incompressibility equations determine the pressure up to an additive constant.

\section{Experiments and Findings} \label{sec:experiments}

In this section, we present the experimental results of CAE-Bench. We first describe the experimental setup in Section~\ref{sec:experimental-setup}. We then compare different models and reasoning-effort settings and discuss API costs in Sections~\ref{sec:medium-comparison}, \ref{sec:re-comparison}, and~\ref{sec:api-cost}, respectively. The generalizability performance is discussed in Section~\ref{sec:generalization-performance}.

\subsection{Experimental Setup} \label{sec:experimental-setup}

We test 8 OpenAI models, including GPT-5.6 Luna and GPT-6 Luna (each at none, low, medium, high, xhigh, and max), as well as GPT-5.6 Sol, GPT-6 Sol, GPT-6.1 Sol, GPT-6 Astra, GPT-6.2 Sol, and GPT-6.2 Astra (each at medium). We also test Gemini 3.8 Flash (medium), Gemini 4 Argon (medium), Gemini 4 Carbon (medium), DeepSeek V4.1 Flash (high), Kimi K2.7 Code (thinking), Qwen3.8-Flash (low, medium, and xhigh), and GLM-5.3-Flash (low, high, and max). In total, we evaluate 15 models across 29 model--effort configurations. We selected these models from those available to our team at the time of testing. Researchers can implement the experiment and submit into \url{https://github.com/AXeonV/CAE-Bench}

The mini-swe-agent 2.4.6~\citep{yang2024sweagent,lieret2026minisweagent} is used for all experiments. For the Gemini models, development and evaluation use local Harbor~\citep{harbor} runs. For the other models, development and evaluation are conducted in separate Hugging Face Jobs.
Agents have 30 to 120 minutes to develop a solver depending on the task.
Each agent is given 8 vCPUs and 24\,GiB of memory. If an agent fails to submit a solver, he receives a score of 0.
More details are discussed in Appendix~\ref{app:experiments}.

We use the 80 computational tasks in RC7 to evaluate the models. Complete results are provided in Appendix~\ref{app:experiments}.

We calculate the score of each task by averaging its case scores, as defined in Section~\ref{sec:assessment}. The final score is the average of the task scores, with equal weight for each task. All scores are reported on a 0-100 scale. We report the results for tasks with varying physical inputs and fixed-problem tasks separately.

\subsection{Model Comparison} \label{sec:medium-comparison}

We compare 15 different models at representative reasoning settings. As shown in Table~\ref{tab:medium-results}, GPT-6 Sol, GPT-6.1 Sol, and GPT-6 Astra achieve full scores on all 24 fixed-problem tasks. However, their average scores on tasks with varying physical inputs are only 39.90, 50.35, and 49.79, respectively. Both GPT-6.2 models also achieve full scores on the fixed-problem tasks, with averages of 49.97 for Sol and 51.57 for Astra on tasks with varying physical inputs. Gemini 4 Carbon likewise achieves full scores on all 24 fixed-problem tasks, with an average of 48.28 on tasks with varying physical inputs. The result shows limited generalization performance. (More discussion in generalization in Section~\ref{sec:generalization-performance}.)

All models perform better on prescribed-PDE tasks than on fluid-mechanics tasks. GPT-6.2 Astra achieves the highest score in both groups, with 87.37 on prescribed-PDE tasks and 40.75 on fluid-mechanics tasks.

\begin{figure*}[!tp]
  \centering
  \includegraphics[width=\linewidth]{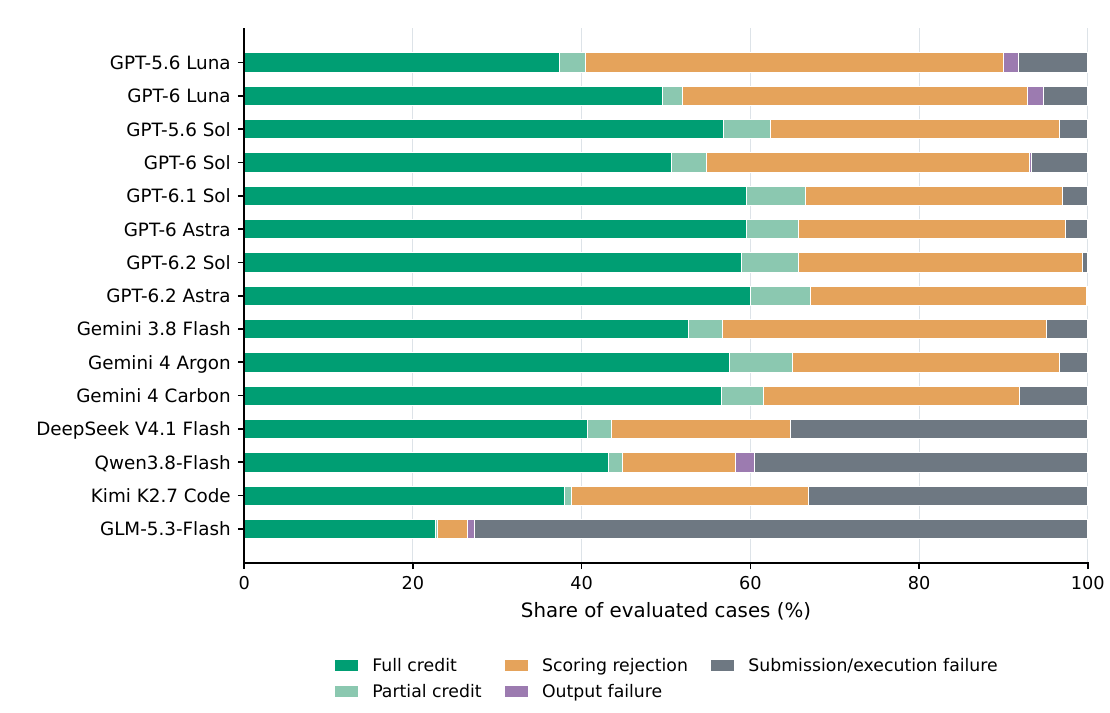}
  \caption{Case outcomes across the 80 computational tasks in RC7.}
  \label{fig:case-outcomes-all-tasks}
\end{figure*}

We further analyze why cases receive 0 score. As shown in Figure~\ref{fig:case-outcomes-all-tasks}, for the 8 OpenAI models, most 0-score cases are caused by scoring rejections rather than execution or output-format failures. For DeepSeek V4.1 Flash, Qwen3.8-Flash, Kimi K2.7 Code, and GLM-5.3-Flash, submission or execution failures occur more frequently.

\begin{table}[!htbp]
  \centering
  \small
  \begin{tabularx}{\linewidth}{@{}Xlccccc@{}}
    \toprule
    Model               & \shortstack{Reasoning\\effort} & \shortstack{Varying physical\\inputs (56)} & \shortstack{PDE\\(13)} & \shortstack{Fluid\\mechanics (43)} & \shortstack{Fixed-problem\\(24)} & Overall \\
    \midrule
    GPT-5.6 Luna        & medium                         & 29.31                                      & 57.39                  & 20.82                              & 68.11                            & 40.95   \\
    GPT-6 Luna          & medium                         & 36.08                                      & 79.13                  & 23.06                              & 75.22                            & 47.82   \\
    GPT-5.6 Sol         & medium                         & 44.37                                      & 87.27                  & 31.40                              & 97.22                            & 60.23   \\
    GPT-6 Sol           & medium                         & 39.90                                      & 72.58                  & 30.02                              & 100.00                           & 57.93   \\
    GPT-6.1 Sol         & medium                         & 50.35                                      & 83.75                  & 40.25                              & 100.00                           & 65.24   \\
    GPT-6 Astra         & medium                         & 49.79                                      & 82.69                  & 39.85                              & 100.00                           & 64.85 \\
    GPT-6.2 Sol         & medium                         & 49.97                                      & 80.78                  & 40.65                              & 100.00                           & 64.98   \\
    GPT-6.2 Astra       & medium                         & 51.57                                      & 87.37                  & 40.75                              & 100.00                           & 66.10   \\
    Gemini 3.8 Flash    & medium                         & 41.06                                      & 75.28                  & 30.71                              & 98.55                            & 58.30   \\
    Gemini 4 Argon      & medium                         & 48.20                                      & 83.91                  & 37.40                              & 95.83                            & 62.49   \\
    Gemini 4 Carbon     & medium                         & 48.28                                      & 78.15                  & 39.26                              & 100.00                           & 63.80   \\
    DeepSeek V4.1 Flash & high                           & 32.73                                      & 54.50                  & 26.15                              & 91.67                            & 50.41   \\
    Qwen3.8-Flash       & medium                         & 27.00                                      & 68.59                  & 14.13                              & 75.00                            & 41.59   \\
    Kimi K2.7 Code      & thinking                       & 24.85                                      & 58.93                  & 14.54                              & 76.29                            & 40.28   \\
    GLM-5.3-Flash       & high                           & 11.92                                      & 30.77                  & 6.23                               & 62.50                            & 27.10   \\
    \bottomrule
  \end{tabularx}
  \caption{Model scores on the 80 computational tasks in RC7. For each model, we average case scores within each task and give equal weight to the evaluated tasks in each column. Scores are on a 0--100 scale.}
  \label{tab:medium-results}
\end{table}

\subsection{Reasoning Effort Comparison} \label{sec:re-comparison}

\begin{figure*}[!tp]
  \centering
  \includegraphics[width=\linewidth,trim={0bp 7bp 0bp 17bp},clip]{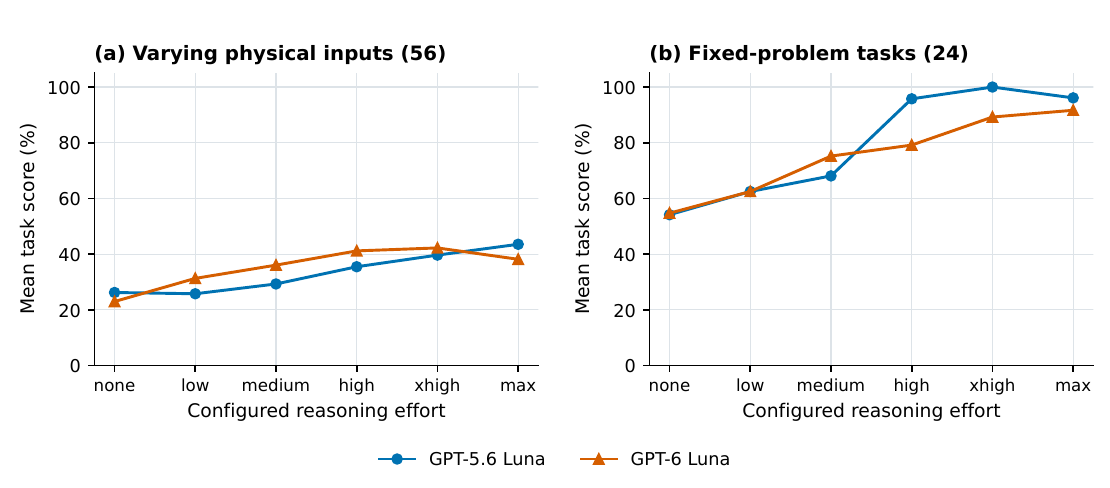}
  \caption{Comparison of reasoning-effort settings for GPT-5.6 Luna and GPT-6 Luna on 80 computational tasks. The tasks include 56 with varying physical inputs and 24 fixed-problem tasks. Each task has equal weight in the average score.}
  \label{fig:reasoning-effort}

  \par\smallskip
  \includegraphics[width=\linewidth]{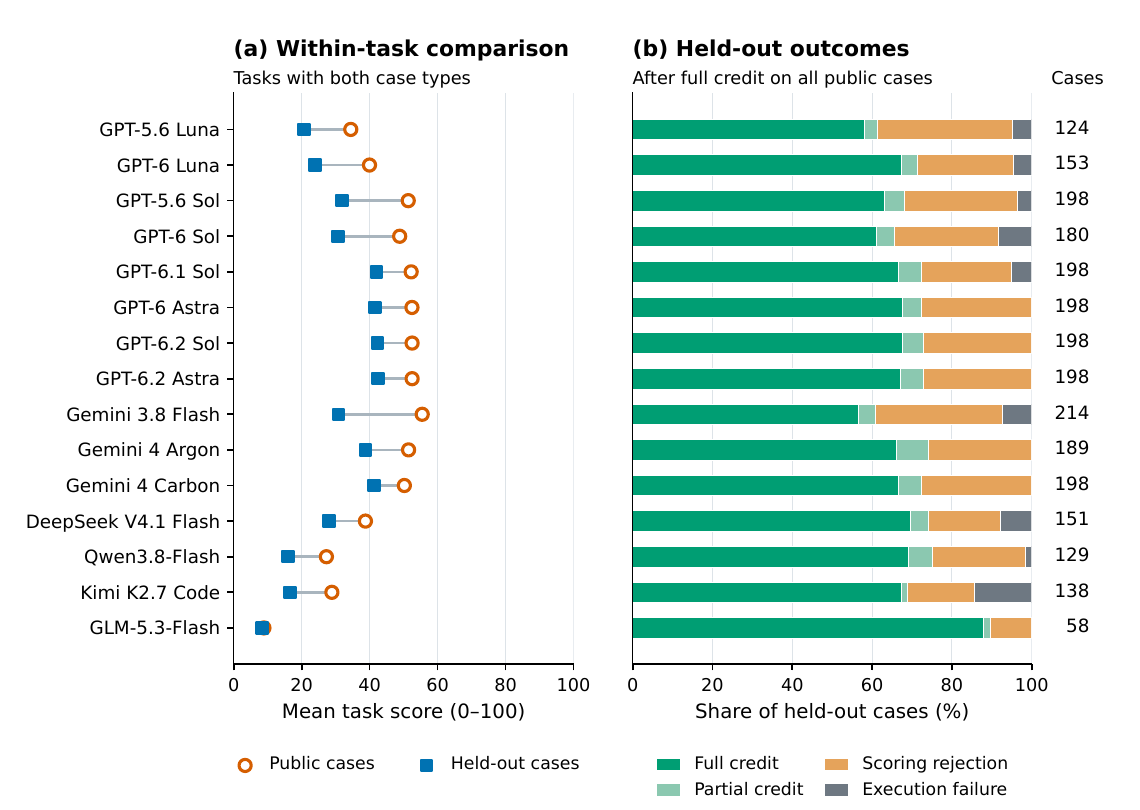}
  \caption{Public and held-out performance on tasks with varying physical inputs. (a) For tasks with both case types, we average public and held-out scores separately within each task and then average across tasks with equal weight. (b) Each bar shows the distribution of held-out case outcomes on tasks where every public case received full score. The numbers on the right give the held-out case count for each bar.}
  \label{fig:heldout-reliability}
\end{figure*}

Figure~\ref{fig:reasoning-effort} shows the scores of GPT-5.6 and GPT-6 Luna at 6 effort settings. The highest effort does not imply a higher score. For example, GPT-6 Luna achieves its highest score on tasks with varying physical inputs at xhigh (42.25), but the score drops to 38.18 at max. At max effort, the model fails to submit a solver for 17 tasks!

\subsection{API Cost and Task Score} \label{sec:api-cost}

Figure~\ref{fig:reward-cost} shows mean task score against mean recorded API cost per task. In general, higher reasoning effort leads to higher costs and higher scores, though this is not always the truth. Under our time limits, higher reasoning effort can increase the time spent generating responses and running tools, leading to lower scores if the agent fails to submit a solver.

\subsection{Generalization Performance} \label{sec:generalization-performance}

As shown in Figure~\ref{fig:heldout-reliability}(a), all models achieve lower scores on held-out cases than on public cases. For example, GPT-6.2 Astra averages 52.49 on public cases and 42.48 on held-out cases.

We also examine a more extreme situation, the tasks where the agents achieve full scores on all public cases. For example, both GPT-6.2 models earn full scores on every public case for 22 tasks each. Their mean held-out scores on their respective sets of tasks are 67.50 for Sol and 67.46 for Astra. Similar results are observed for all models, as shown in Table~\ref{tab:heldout-losses}. Therefore, the generalization performance of all models is limited.

Figure~\ref{fig:heldout-reliability}(b) shows how agents failed on held-out cases: most zero-score held-out cases are caused by scoring rejections while execution failure accounts for the rest.

\section{Limitations} \label{sec:evidence-limits}

CAE-Bench mainly covers prescribed-PDE and fluid-mechanics tasks. The scores only reflect performance on the selected cases and do not guarantee that the developed solvers can handle all possible inputs or other CAE problems.

The reference results are collected from analytical solutions, existing datasets, and simulation software. However, not all reference results have undergone independent numerical verification (see \citealp{oberkampf2002verification} for verification methods).

Although held-out cases are not provided during development, their source data may have appeared in model training data. Moreover, public and held-out cases differ in physical inputs and coverage. Therefore, the score differences cannot be attributed solely to withholding the cases. How benchmark scores should be interpreted and whether agent performance transfers beyond the benchmark are broader evaluation questions~\citep{reuel2024betterbench,patel2026predictive}.

More details are provided in Appendix~\ref{app:experiments}.

\section{Conclusion} \label{sec:conclusion}

We introduced CAE-Bench, a benchmark for evaluating the ability of AI agents to develop reusable CAE solvers. Unlike existing benchmarks that focus on fixed problems, CAE-Bench evaluates the generalization of AI-developed solvers across unseen physical inputs. Our experiments show that current AI agents are not capable of developing reusable CAE solvers without human involvement. Even when an AI-developed solver performs perfectly on cases used for development, it may fail on unseen cases. We hope CAE-Bench will help researchers develop AI agents capable of building reliable and reusable CAE solvers.

\section*{Acknowledgement}

We thank the EIT High Performance Computing Platform for providing computational resources for this project.
We also thank Wei Zhang of Southern University of Science and Technology for access to the Titan workstation. Many computations in our study would have been difficult without access to the workstation.
\bibliography{references}
\clearpage
\appendix
\section{Data Format and Public Tools}
\label{app:contract}

Each task contains \texttt{instruction.md}, \texttt{DATA.md}, and the shared \texttt{CONTRACT.md} as the test materials provided to agents.

\texttt{instruction.md} is the overview information in the problem. It also gives the input data range. \texttt{CONTRACT.md} is the general data protocol for data storage and referencing. It serves as a common set of rules suitable for all problems. \texttt{DATA.md} defines the comprehensive rules for certain physical fields and constraints. Any specific rule not covered by \texttt{CONTRACT.md} is declared in \texttt{DATA.md}.

When the problem packages are being assembled, the \texttt{problem.json} will be generated for each case. The solver reads the input from \texttt{problem.json}. This file specifies the physical model, parameters and conditions. The information in \texttt{fields} and \texttt{outputs} describes the input fields and output requirements. Each term of declaration gives the details of HDF5 data such as dataset name, physical quantity, unit, data type, and component labels and shape. It may also specify the location and time axis, or explain whether the data are point values, averages or integrals. Conditions can be given as constants with units or refer to data stored in the input fields in an HDF5 file.

The input from an HDF5 file stores the concrete field array, entity id number and time value in the dataset. If there is a need to use a mesh, \texttt{problem.json} will also provide a Gmsh \texttt{.msh} file in MSH 4.1 ASCII format. The mesh file stores the nodes and edges information which is connected to the entity id number. Otherwise, the locations of this entity id number will be stored in the HDF5 file directly.

The axes of a field are in order: time axes, entity axes and component axes (declared in \texttt{problem.json}). Every position in the entity axes is aligned to the entity id number. The solver needs to write the output field to a separate HDF5 file and strictly ensure the order.

\subsection{A Released Heat-Conduction Example}
\label{app:heat-example}

The public samples of the problem \texttt{heat-conduction-steady-2d} are in \texttt{samples/01}. The problem is designed to solve steady-state heat conduction on a unit square with no heat source. The thermal conductivity is 1\,W/(m\,K). The temperature of 4 boundaries is 4\,K. The mesh includes 9 nodes and 8 triangle units. Full information about the problem is available in \path{supplement/heat-conduction-steady-2d-agent.zip}.

The sample \path{input/problem.json} uses \texttt{mesh.msh} and \texttt{input.h5}. Here are the condition on the left boundary and the output requirement.

\begin{lstlisting}
{
  "conditions": [
    {"role": "boundary", "variable": "temperature",
     "location": "left", "operator": "dirichlet",
     "data": {"coefficients": {"field": "left_coefficients"}}}
  ],
  "outputs": {
    "temperature": {
      "array": "temperature", "quantity": "temperature",
      "location": "nodes", "time_axis": "steady",
      "dtype": "float64", "kind": "scalar",
      "component_shape": [1], "components": ["value"],
      "unit": "K", "value_kind": "point"
    }
  }
}
\end{lstlisting}

The boundary condition uses \texttt{left} to refer to the \texttt{left\_ids} dataset. It  stores the element tags of the left boundary. The boundary condition also uses \texttt{left\_coefficients} and gives the name to the \texttt{input.h5} dataset accordingly. The dataset stores 5 temperature polynomial coefficients for each boundary in ascending power order. The polynomial in this case is always 4\,K on the left boundary.

The output requirement uses \texttt{nodes} and \texttt{steady} to refer to \texttt{node\_ids} and \texttt{times} in the \texttt{input.h5} dataset. The shapes of the 3 datasets are  $(9,)$, $(1,)$, and $(1,9,1)$, respectively. \texttt{temperature[0,i,0]} gives the temperature in the time value 0 of \texttt{node\_ids[i]}. This format is demonstrated by \path{output/result.h5} using random temperature values.

\subsection{Using the Data Tools}
\label{app:data-tools}

The solver uses \texttt{--input} to receive the path of \texttt{problem.json} and \texttt{--output} to receive the path of the HDF5 file it needs to write. The solver can also use SDKs provided by us to read and write. \texttt{load\_problem} helps read and check the input. \texttt{read\_field} helps read a dataset from the input HDF5 file according to the problem \texttt{problem.json}. \texttt{write\_result} helps align and write the calculation output into the arrays. It also helps check the structure of the results under the rules of our \texttt{DATA.md} and \texttt{CONTRACT.md}.

The package also provides a \emph{validator} to make structural checks. The \emph{validator} is designed to reduce the burden of reading and writing data under our contract. We do not want the agents to have problems on understanding the dataset input or make errors on the output structure instead of the problem itself. The \emph{validator} checks formatting, references, and declaration constraints. It does not verify or correct physical results.

Run the following commands in a Python environment to check the released example.

\begin{lstlisting}
python -m pip install .
cd qbank/01-prescribed-pdes/1.1-elliptic/\
heat-conduction-steady-2d/samples/01
python -m caeoj validate-input input/problem.json
python -m caeoj validate-output --input input/problem.json \
  --output output/result.h5
\end{lstlisting}

\texttt{validate-input} checks the JSON declaration and mesh and HDF5 reference of \texttt{problem.json}. \texttt{validate-output} checks whether the data type and shape meet the requirements of the entity number id and time value.

\texttt{python -m caeoj grade} performs numerical grading. It grades using input cases and grading criteria \texttt{grading.json} on the solver.

\section{Reference and Release Records}
\label{app:reference-records}

We list supplement files used in our experiments to help evaluate and reproduce our results.

\subsection{Release Identity and Access}
\label{app:release-access}

The tasks and test cases used in our experiments are available in RC7 (\url{https://github.com/AXeonV/CAE-Bench/releases/tag/caeoj-qbank-20261029-rc7}). The source repository is \url{https://github.com/AXeonV/CAE-Bench}. The supplement describes how to download and verify the release files.

The full dataset is available in the CAE-Bench dataset repository on Hugging Face (\url{https://huggingface.co/datasets/jimzhouzzy2006/CAE-Bench}). Table~\ref{tab:supplement-inventory} shows the detailed path and records.

\begin{table}[!htbp]
  \centering
  \small
  \begin{tabularx}{\linewidth}{@{}p{0.43\linewidth}X@{}}
    \toprule
    File under \texttt{supplement/}            & Contents                     \\
    \midrule
    \path{reference_records.json}              & 
    Reference data index and source                                           \\
    \path{evidence/}                           & 
    Record of reference data making process                                   \\
    \path{experiment_results.json}             & 
    Experiment results and model configurations                               \\
    \path{heldout_losses.json}                 & 
    Held-out scores (tasks with full credit on all public cases)              \\
    \path{score_group_diagnostics.json}        & 
    Report of each group (errors and checks)                                  \\
    \path{score_sensitivity.json}              & 
    Result under different scoring rules                                      \\
    \path{evaluation_manifest.json}            & 
    Evaluation file mappings, hashes, and resource limits                     \\
    \path{evaluation_overlay.zip}              & 
    Modified and added evaluation files                                       \\
    \path{evaluation_runtime/}                 & 
    Evaluation runtime source code                                            \\
    \path{heat-conduction-steady-2d-agent.zip} & 
    Public heat-conduction task package used in Appendix~\ref{app:contract}.  \\
    \path{resource_locations.json}             & 
    Record of repository and dataset location                                 \\
    \path{MANIFEST.json}                       & 
    File hash and size                                                        \\
    \bottomrule
  \end{tabularx}
  \caption{Supplementary files provided with CAE-Bench.}
  \label{tab:supplement-inventory}
\end{table}

\subsection{Public Packages and Evaluation Files}
\label{app:public-private-files}

Each of our \texttt{public/<task>-agent.zip} includes our data format and public tools defined in Appendix~\ref{app:contract}. \texttt{qbank.zip} contains reference data, grading criteria and some tools to create these questions. The sample data in \texttt{public/<task>-agent.zip} are not extracted from the reference data. The reference data are not visible to agents in experiments. To ensure the rigor of experiments, we halt the agents and grade their solver outside the environment independently.

\subsection{Reference Sources and Evidence}
\label{app:reference-evidence}

\path{supplement/reference_records.json} records 82 tasks with 647 evaluation cases and 96 format examples. Reference data are associated with their sources and related information in the records above. These records also include how these data were generated and evaluated by us.

For the reference results, all heat-conduction cases use analytical solutions. The 2-dimensional steady-flow uses 8 analytical solutions and 2 Lethe~\citep{blais2020lethe} calculations. The 3-dimensional steady-flow uses 10 Elmer~\citep{ruokolainenelmer} generations. Others use the calculation results from OpenFOAM~\citep{jasak2009openfoam}, Elmer, SPARTA~\citep{plimpton2019sparta}, PyClaw~\citep{ketcheson2012pyclaw}, Thetis~\citep{karna2018thetis}, and formula evaluation. We also use NGSolve~\citep{schoberl2014ngsolve} and Dedalus~\citep{burns2020dedalus} for some fixed-problem cases. \path{supplement/reference_records.json} records the settings used to generate these results and explains how we convert the source fields to the requested values.

For the tasks adapted by PDEBench, some of them are directly extracted from the dataset while others are generated by us using the well-known solvers mentioned above. We do not guarantee that all of the reference data are credible enough. So we record the question creating process as detailed as possible in the related files.

\subsection{Reconstruction Procedure and Limits}
\label{app:reconstruction}

Rename the restored directory to \texttt{rc7}. Then run the command in a Python environment to check file integrity.

\begin{lstlisting}
python scripts/rebuild_evaluation_snapshot.py \
  --release-dir rc7 \
  --manifest supplement/evaluation_manifest.json \
  --overlay supplement/evaluation_overlay.zip --verify-only
\end{lstlisting}

\path{supplement/evaluation_REPRODUCE.md} explains how to extract a single task or the whole to a new directory and how to grade a new submitted solver. When grading, use the solver to generate a \texttt{result.h5} from a certain case. Then use the grading command \texttt{python -m caeoj grade} to evaluate the result. The command itself does not check if the solver exceeds its runtime limits by \texttt{judge-config.json}.

\subsection{Sensitivity to Scoring Choices} \label{sec:score-sensitivity}

We test how different scoring rules affect the model comparison. We recompute the scores of the 15 model configurations in Table~\ref{tab:medium-results} on tasks with varying physical inputs using the saved numerical errors, without rerunning the solvers. All other scoring settings remain unchanged, and cases that failed mandatory checks still receive 0 score.

Table~\ref{tab:score-sensitivity} compares 2 changes to the scoring rule. First, we change the full-score and 0-score thresholds by $\pm20\%$ from the original $(0.05,0.10)$. Second, we replace RMS group aggregation with the maximum group error while keeping the original thresholds.

\begin{table}[!htbp]
  \centering
  \small
  \begin{tabularx}{\linewidth}{@{}Xrrrr@{}}
    \toprule
    Model               & \shortstack{RMS\\4\%/8\%} & \shortstack{RMS\\5\%/10\%} & \shortstack{RMS\\6\%/12\%} & \shortstack{Max\\5\%/10\%} \\
    \midrule
    GPT-5.6 Luna        & 28.52                     & \textbf{29.31}             & 29.99                      & 27.88                      \\
    GPT-6 Luna          & 35.41                     & \textbf{36.08}             & 36.60                      & 34.16                      \\
    GPT-5.6 Sol         & 42.56                     & \textbf{44.37}             & 46.46                      & 38.83                      \\
    GPT-6 Sol           & 38.16                     & \textbf{39.90}             & 41.22                      & 34.43                      \\
    GPT-6.1 Sol         & 47.26                     & \textbf{50.35}             & 52.70                      & 43.01                      \\
    GPT-6 Astra         & 46.93                     & \textbf{49.79}             & 52.05                      & 42.77 \\
    GPT-6.2 Sol & 47.03 & \textbf{49.97} & 52.32 & 42.38 \\
    GPT-6.2 Astra & 48.53 & \textbf{51.57} & 53.86 & 43.86 \\
    Gemini 3.8 Flash    & 39.65                     & \textbf{41.06}            & 42.40                      & 36.72                      \\
    Gemini 4 Argon      & 45.48                     & \textbf{48.20}            & 50.61                      & 40.63                      \\
    Gemini 4 Carbon     & 45.89                     & \textbf{48.28}            & 50.36                      & 42.10                      \\
    DeepSeek V4.1 Flash & 31.53                     & \textbf{32.73}             & 33.57                      & 28.60                      \\
    Qwen3.8-Flash       & 26.31                     & \textbf{27.00}             & 27.59                      & 25.54                      \\
    Kimi K2.7 Code      & 24.63                     & \textbf{24.85}             & 25.04                      & 23.96                      \\
    GLM-5.3-Flash       & 11.85                     & \textbf{11.92}             & 11.97                      & 11.85                      \\
    \bottomrule
  \end{tabularx}
  \caption{Score sensitivity on tasks with varying physical inputs for the 15 model configurations in Table~\ref{tab:medium-results}. Each entry averages the task scores with equal task weights. Scores use a 0--100 scale. We recompute the scores from stored group errors for the same submitted programs. Column headings give the full-score and zero-score thresholds, respectively. The bold column uses the official RMS 5\%/10\% scoring rule. Failures of execution, output-format checks, and exact-value checks retain 0 score.}
  \label{tab:score-sensitivity}
\end{table}

The 4\%/8\% RMS threshold preserves the ranking, while the 6\%/12\% RMS threshold swaps only Gemini 4 Carbon and Gemini 4 Argon. Using the maximum group error lowers the scores and swaps GPT-6.2 Sol and GPT-6 Astra. GPT-6.2 Astra remains first under all four scoring rules.

Under the original RMS scoring rule, a case can receive full score even if some groups have NRMSE above 5\%. This occurs in 231 of the 4,359 full-score case outcomes (5.3\%).

\FloatBarrier

\section{Additional Experimental Results}
\label{app:experiments}

The model comparison in Section~\ref{sec:medium-comparison} uses one reasoning setting per model. Here we give the results for all 29 model--effort configurations and describe how we ran the experiments and calculated costs.

\subsection{Scores and Task Coverage}
\label{app:comparison-records}

Table~\ref{tab:all-cases} shows the score and API cost under a certain effort of one model. Each row represents a valid report of tasks. The weight of each task in one task group is equal. If the evaluation process is done but the agent gives no submission, the score will be 0.

Qwen3.8-Flash has 79 reports at medium effort, with \path{resolved-statistics-3d} missing. Every other configuration has reports for all 80 tasks.

\path{supplement/experiment_results.json} gives the detailed results per task and case.

\begin{table}[!t]
  \centering
  \small
  \renewcommand{\arraystretch}{0.94}
  \begin{tabularx}{\linewidth}{@{}Xlcccccc@{}}
    \toprule
                        &          & \multicolumn{2}{c}{Varying physical inputs} & \multicolumn{2}{c}{Fixed problems} &        &                              \\
    \cmidrule(lr){3-4}\cmidrule(lr){5-6}
    Model               & Effort   & Score                                       & USD/task                           & Score  & USD/task & Overall & Reports \\
    \midrule
    GPT-5.6 Luna        & none     & 26.25                                       & 0.0193                             & 54.17  & 0.0110   & 34.63   & 80/80   \\
                        & low      & 25.80                                       & 0.0133                             & 62.50  & 0.0103   & 36.81   & 80/80   \\
                        & medium   & 29.31                                       & 0.0216                             & 68.11  & 0.0154   & 40.95   & 80/80   \\
                        & high     & 35.49                                       & 0.0666                             & 95.77  & 0.0352   & 53.58   & 80/80   \\
                        & xhigh    & 39.67                                       & 0.1346                             & 100.00 & 0.0545   & 57.77   & 80/80   \\
                        & max      & 43.56                                       & 0.2668                             & 96.11  & 0.0791   & 59.32   & 80/80   \\
    \addlinespace[1pt]
    GPT-6 Luna          & none     & 23.00                                       & 0.0136                             & 54.73  & 0.0071   & 32.52   & 80/80   \\
                        & low      & 31.32                                       & 0.0060                             & 62.50  & 0.0045   & 40.67   & 80/80   \\
                        & medium   & 36.08                                       & 0.0188                             & 75.22  & 0.0088   & 47.82   & 80/80   \\
                        & high     & 41.19                                       & 0.0356                             & 79.17  & 0.0153   & 52.58   & 80/80   \\
                        & xhigh    & 42.25                                       & 0.0570                             & 89.24  & 0.0185   & 56.35   & 80/80   \\
                        & max      & 38.18                                       & 0.0785                             & 91.67  & 0.0344   & 54.22   & 80/80   \\
    \addlinespace[1pt]
    GPT-5.6 Sol         & medium   & 44.37                                       & 0.7407                             & 97.22  & 0.4077   & 60.23   & 80/80   \\
    \addlinespace[1pt]
    GPT-6 Sol           & medium   & 39.90                                       & 0.2523                             & 100.00 & 0.1411   & 57.93   & 80/80   \\
    \addlinespace[1pt]
    GPT-6.1 Sol         & medium   & 50.35                                       & 0.5759                             & 100.00 & 0.1495   & 65.24   & 80/80   \\
    \addlinespace[1pt]
    GPT-6 Astra         & medium   & 49.79                                       & 4.2573                             & 100.00 & 0.9462   & 64.85   & 80/80 \\
    \addlinespace[1pt]
    GPT-6.2 Sol & medium & 49.97 & 0.7795 & 100.00 & 0.1962 & 64.98 & 80/80 \\
    \addlinespace[1pt]
    GPT-6.2 Astra & medium & 51.57 & 2.5351 & 100.00 & 0.3084 & 66.10 & 80/80 \\
    \addlinespace[1pt]
    Gemini 3.8 Flash    & medium   & 41.06                                       & 1.3444                             & 98.55  & 0.7194   & 58.30   & 80/80   \\
    \addlinespace[1pt]
    Gemini 4 Argon      & medium   & 48.20                                       & 3.7907                             & 95.83  & 0.4144   & 62.49   & 80/80   \\
    \addlinespace[1pt]
    Gemini 4 Carbon     & medium   & 48.28                                       & 5.8800                             & 100.00 & 0.8316   & 63.80   & 80/80   \\
    \addlinespace[1pt]
    DeepSeek V4.1 Flash & high     & 32.73                                       & 0.2160                             & 91.67  & 0.1072   & 50.41   & 80/80   \\
    \addlinespace[1pt]
    Qwen3.8-Flash       & low      & 30.16                                       & 0.1798                             & 70.83  & 0.0520   & 42.36   & 80/80   \\
                        & medium   & 27.00                                       & 0.1689                             & 75.00  & 0.0664   & 41.59   & 79/80   \\
                        & xhigh    & 16.83                                       & 0.1378                             & 66.67  & 0.1198   & 31.78   & 80/80   \\
    \addlinespace[1pt]
    Kimi K2.7 Code      & thinking & 24.85                                       & 1.4397                             & 76.29  & 0.5807   & 40.28   & 80/80   \\
    \addlinespace[1pt]
    GLM-5.3-Flash       & low      & 21.38                                       & 0.3434                             & 73.55  & 0.1108   & 37.03   & 80/80   \\
                        & high     & 11.92                                       & 0.1705                             & 62.50  & 0.1245   & 27.10   & 80/80   \\
                        & max      & 16.39                                       & 0.2463                             & 66.67  & 0.1307   & 31.47   & 80/80   \\
    \bottomrule
  \end{tabularx}
  \caption{Scores and mean recorded API costs on RC7. Scores are on a 0--100 scale. The last column gives the number of valid task reports out of 80.}
  \label{tab:all-cases}
\end{table}

\subsection{Development and Evaluation Conditions}
\label{app:development-budgets}
\label{app:evaluation-resources}

In our experiments, the agent develops the solver in a public task workspace. It can use a shell to read and write files, run programs, and access the network.

For the Gemini models, development and evaluation use local Harbor runs. For the other models, development and evaluation run in separate Hugging Face Jobs. The jobs use 8 vCPUs and a 24\,GiB memory allowance. The evaluator runs each case in a separate process without network access. The solver is required to produce \texttt{result.h5} per task in between 4\,s and 40\,min. The time limit depends on each task.

During grading, the solver output \texttt{result.h5} is compared to reference results. For most tasks, the grader has 60\,s to judge per case. For the 3D compressible Navier--Stokes task, the solver has 30\,min to produce \texttt{result.h5} and the grader has 2\,min to judge.

\subsection{Held-out Performance After Full Public Score}

For each model, Table~\ref{tab:heldout-losses} gives the number of tasks with full score on every public case and the mean score on their held-out cases. Only tasks with both public cases and held-out cases are included. Each task has equal weight.

\begin{table}[!t]
  \centering
  \small
  \begin{tabularx}{\linewidth}{@{}Xcc@{}}
    \toprule
    Model               & \shortstack{Tasks with full score\\on all public cases} & \shortstack{Mean held-out\\score (0--100)} \\
    \midrule
    GPT-5.6 Luna        & 15                                                      & 57.35                                      \\
    GPT-6 Luna          & 17                                                      & 60.98                                      \\
    GPT-5.6 Sol         & 22                                                      & 59.16                                      \\
    GPT-6 Sol           & 20                                                      & 60.13                                      \\
    GPT-6.1 Sol         & 22                                                      & 66.97                                      \\
    GPT-6 Astra         & 22                                                      & 67.44 \\
    GPT-6.2 Sol & 22 & 67.50 \\
    GPT-6.2 Astra & 22 & 67.46 \\
    Gemini 3.8 Flash    & 24                                                      & 53.91                                       \\
    Gemini 4 Argon      & 21                                                      & 65.56                                       \\
    Gemini 4 Carbon     & 22                                                      & 66.80                                       \\
    DeepSeek V4.1 Flash & 16                                                      & 68.30                                      \\
    Qwen3.8-Flash       & 12                                                      & 57.75                                      \\
    Kimi K2.7 Code      & 13                                                      & 55.56                                      \\
    GLM-5.3-Flash       & 4                                                       & 81.78                                      \\
    \bottomrule
  \end{tabularx}
  \caption{Mean held-out scores after full score on all public cases for the cases in Table~\ref{tab:medium-results}.}
  \label{tab:heldout-losses}
\end{table}

\subsection{API Cost Estimates}
\label{app:cost-estimates}

\begin{figure*}[!tp]
  \centering
  \includegraphics[width=\linewidth]{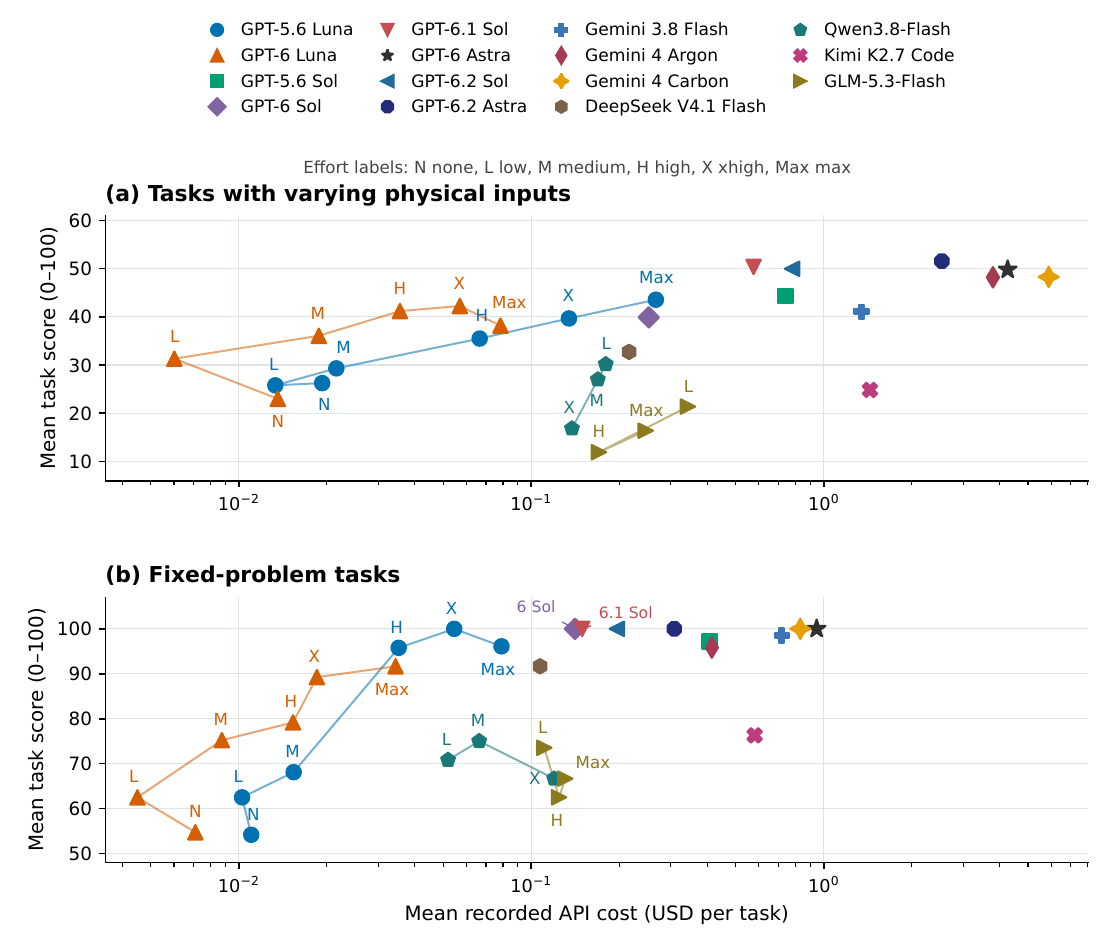}
  \caption{Recorded API costs are in USD per task on a logarithmic scale.}
  \label{fig:reward-cost}
\end{figure*}

For each submission, we estimate the API cost by the recorded token counts. We use the provider's rates for input and output tokens, including the rate for cached input where applicable. In our experiments, a run can be interrupted by the provider's rate limit and be resumed. When a run resumes, we add the cost of the new requests to the cost recorded before it resumed. In Figure~\ref{fig:reward-cost}, we average scores and costs over the same tasks for each configuration.

We price DeepSeek usage at the peak rate. We convert CNY to USD at USD\,0.150310793 per CNY using the reference rates for October 30, 2026 from the European Central Bank~\citep{ecb2026exchangerates}.

\FloatBarrier

% Keep the heading on the same page as the first supplementary figure.
\begin{figure*}[!tp]
  \section{Supplementary Figures}
  \label{app:supplementary-figures}
  \centering
  \includegraphics[width=\linewidth]{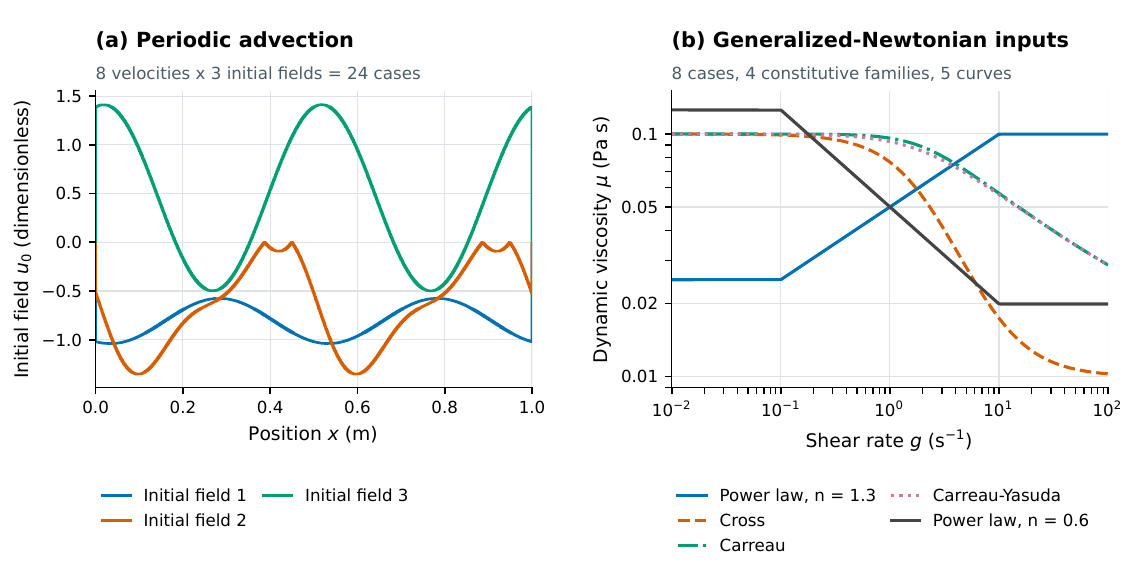}
  \caption{Input variations for periodic advection and generalized-Newtonian flow tasks.}
  \label{fig:case-variation}
\end{figure*}

\end{document}